\pdfoutput=1

\documentclass[11pt]{article}

\usepackage[preprint]{acl}

\usepackage{times}
\usepackage{latexsym}

\usepackage[T1]{fontenc}

\usepackage[utf8]{inputenc}

\usepackage{microtype}

\usepackage{inconsolata}

\usepackage{graphicx}

\usepackage[dvipsnames]{xcolor}

\usepackage{caption}
\usepackage[english]{babel}
\title{Right-Wing Rock or Just Rock? A Computational Linguistic Analysis of Frei.Wild}

\author{
\textbf{Carlotta Schneeberger}$^1$, \textbf{Kevin Tang}$^{2,3}$,\\
Faculty of Arts and Humanities, Heinrich Heine University Düsseldorf \\
$^1$Institute of Linguistics \hspace{1em} $^2$Department of English Language and Linguistics,\\
$^3$Department of Linguistics, College of Liberal Arts and Sciences, University of Florida\\
\texttt{\{Carlotta.Schneeberger, kevin.tang\}@uni-duesseldorf.de}
}

\begin{document}
\maketitle
\begin{abstract}

Rechtsrock is a subgenre of rock music that spreads right-wing ideology, often instrumentalized to recruit adolescents into the radical scene. Monitoring institutions counteract this by manually examining and, in some cases, banning extremist content; however, there are border cases that evade regulation.
We present a study aimed at determining whether such a case, the band \textit{Frei.Wild}, should be classified as politically right-leaning or as part of the general German rock genre. We sampled a German rock dataset and created a corpus for right-wing rock to use as reference in this analysis and found that we can confirm the intuitions from previous investigations that \textit{Frei.Wild} successfully maintains an ambiguity with regard to their political affiliation. However, the tendency is towards the right-wing spectrum. Lexical analyses reveal nationalistic narratives and two high-performing classifiers (up to 97\% ROC-AUC score) label more than half of their songs as right-wing extremist.
Our analysis provides insight into how computational methods can improve the process of identifying right-wing extremist tendencies in music, especially in borderline cases like \textit{Frei.Wild}. The code and data are made available for future research.
\end{abstract}

\section{Introduction}
\label{sec:intro}
Rechtsrock, which translates to ``right-wing rock,'' is a subgenre of rock music characterized by its nationalistic, racist, and xenophobic lyrics. The music often romanticizes themes of racial purity and national pride, and promotes historical revisionism. It has been argued to act as a ``gateway drug'' or ``recruitment tool'' into the radical scene and for spreading extremist ideologies \citep{Moeller2020,BfV2020RechtsextremistischeErlebniswelt,vsb2024}.
Just as the Nazi regime's use of music to foster community and their use of emotions to mediate its ideology, right-wing rock music harnesses emotional appeal to promote the neo-Nazi worldview \citep[p. 18]{buechner2018weltbuergertum}.

Qualitative studies have analyzed Rechtsrock lyrics for their themes and strategies \citep{dornbusch2002,naumann2009}, identifying among other things `Heimat' (homeland) in association with `Blut, Ehre, Stolz, Vaterland, Reich' (blood, honor, pride, fatherland, empire) as a relevant underlying concept in Rechtsrock \citep[383]{hofig2000}.
In light of these studies, criticism of border cases like the former skinhead band \textit{Böhse Onkelz} \citep{Farin2005} or the `Heimatrock' (homeland rock) band \textit{Frei.Wild} grew \citep{Fischer2022,buchner2019,balzer2019}. Since their debut in 2002 , \textit{Frei.Wild}, and especially their singer Philipp Burger, who used to be a neonazi-skinhead and part of the right-wing extremist band \textit{Kaiserjäger}, have been criticized for nationalistic lyrics. \textit{Frei.Wild} had to distance themselves multiple times from their right-wing extremist fans \citep[p. 167]{Hindrichs2014} that found common ground in songs about the need to defend their homeland and values, next to relativizing the Shoah \citep[pp. 166, 175]{Hindrichs2014}. Still, labeling their music as `youth-endangering content' was rejected by the \citet[p. 20, `Federal Agency for Child- and Youth Media Protection']{bpjm}.

This case-study addresses these mixed findings from qualitative studies and institutional judgments using computational linguistic methods, to give a more data-driven perspective of the similarities between \textit{Frei.Wild} and the Rechtsrock genre and thereby tracking right-wing tendencies in mainstream music. With a direct comparison of their lyrics, we also sidestep the problem of needing a precise definition of Rechtsrock that e.g. investigative journalist Thomas Kuban was missing \citep[p. 176]{Hindrichs2014} when he categorized \textit{Frei.Wild} as such \citep{stz2013,kuban2012}.

The paper is organized as follows. In Section~\ref{sec:background}, we review qualitative studies on right-wing extremist lyrics as well as computational approaches applied to right-wing extremist data. Section~\ref{sec:data} describes the datasets, and Section~\ref{sec:methods} details our methodology. In Sections~\ref{sec:lexical} and ~\ref{sec:classification}, we present our lexical analysis and classification experiments, respectively. Finally, Section~\ref{sec:discussion} discusses the overall results, and Section~\ref{sec:conclusion} concludes the paper and outlines directions for future work. The code and data are publicly available at \url{https://zenodo.org/records/22676753}.

\section{Background}
\label{sec:background}

A prominent approach in studies that look at Rechtsrock/right-wing extremist lyrics is to do a qualitative analysis of one or more songs, to investigate which facets of Nazi-ideology are conveyed in what way.
\citet{Moeller2020}, for example, focus on the bands \textit{Landser} and \textit{Sturmwehr} that are very popular in the Rechtsrock scene and are frequently cited in court files of German prosecutors and in previous studies \citep{naumann2009}. They conducted a sequential text analysis of two songs (`Sturmführer' and `Bis hier her')\footnote{A translation of the song titles mentioned in the current paper is provided in Appendix~\ref{sec:appendixsong}.}, identifying central themes and references by doing a close reading. Their analysis highlights narratives of enemy construction and oppression, a self-image of superiority, emphases on loyalty and resistance, and the use of violent imagery. To contextualize these findings, they compare the songs with two other songs from the Salafi jihadist scene.

Similar methods have been applied to the lyrics of \textit{Frei.Wild}. \citet{alt2020} aims to illustrate the nationalization of German popular music by analyzing works of three German artists, including \textit{Frei.Wild}. This study investigates how New Right ideology is introduced into the musical mainstream. He concludes that songs such as `Wahre Werte', `Südtirol', `Antiwillkommen', `Gutmenschen und Moralapostel', and `Schlagzeile groß Hirn zu klein' articulate a patriotic and nationalistic view of homeland, culture, and tradition, while constructing a narrative of `us against them'. As \citet[p. 302]{kuban2012} notes, such discourse contributes to the popularization of ``nationalism and anti-antifascism''.

To the best of our knowledge, no computational study to date has specifically focused on Rechtsrock. However, there are computational studies that investigate right-wing extremist data. There even is a broader interest in German right-wing research in the area of NLP, for instance, \citet{stede2025} gathered a corpus of the German right-wing populist party \textit{Alternative für Deutschland} speaking on climate change. Lexical and classification analyses are applied to detect populism and emotions, revealing that in contrast to the other German parliament parties they use a higher level of populist language and negative emotions (especially anger) around climate change discourse.
\citet{hartung2017} developed a machine learning approach intended to support manual content moderation efforts by identifying right-wing extremist content in German Twitter (now X) profiles. They framed this task as a binary classification problem and utilized TF-IDF vectors to represent the data, incorporating four distinct feature sets: lexical (a bag-of-words (BOW) frequency profile of all tokens), the emotion conveyed by the tweet, the user's arguments (for or against specific topics), and social identity. The resulting classifier achieved high recall on right-wing extremist profiles, with experiments demonstrating that all features were reliable predictors of political orientation, and notably, the lexical cues exhibited the strongest individual contribution, remaining unsurpassed by any combination of the other features.
Similarly, \citet{dragos2023} compared different binary classification methods for extremist content in French. They used three feature-based models developed with Support Vector Machine (SVM), utilizing data representations based on TF-IDF, BOW, and four sets of features: surface, domain specific (hate-specific words), polarity of words (positive, negative and neutral), and linguistic features (specific verbs). These were contrasted with CamemBERT, a pre-trained language model for French. The models were tested in two experiments: one on an initial imbalanced data set and a second on a re-balanced set using oversampling. They found that, generally, CamemBERT performed best on Accuracy and Precision across both experiments. However, in the second experiment (with re-balanced data), the model using SVM + TF-IDF vectors achieved similar results for Accuracy and Precision, but provided a better overall F-score, thus outperforming CamemBERT on that specific metric.

\section{The Data}
\label{sec:data}
Our study employs three distinct corpora: a corpus of songs by the band \textit{Frei.Wild}, and two reference corpora -- a corpus of right-wing extremist rock music (Rechtsrock) and a corpus of general German rock. The Rechtsrock corpus, due to restricted availability, set a limit on size, while the \textit{Frei.Wild} corpus restricted the timeframe of the other corpora. Table~\ref{tab:corpus_stats} provides a summary of the three corpora, including the number of songs and their average length, number of bands, and temporal distribution.
\begin{table}[ht]
\centering
\resizebox{\linewidth}{!}{%
\begin{tabular}{|l|c|c|c|c|c|c|}
\hline
\textbf{Corpus} & \textbf{Songs} & \textbf{Bands} & \textbf{1990s} & \textbf{2000s} & \textbf{2010s} & \textbf{avg. song len} \\
\hline
Rechtsrock & 99 & 9 & 22 & 57 & 20 & 224 words \\
\hline
German Rock & 99 & 9 & 22 & 55 & 22 & 237 words\\
\hline
\textit{Frei.Wild} & 47 & 1 & 0 & 25 & 22 & 264 words \\
\hline
\textit{Frei.Wild} (full) & 303 & 1 & 0 & 68 & 235 & 243 words\\
\hline
\end{tabular}
}
\caption{Summary statistics of the three corpora.}
\label{tab:corpus_stats}
\end{table}

\subsection{The Reference Corpus for Rechtsrock}
As a reference for Rechtsrock, the Rechtsrock corpus was compiled by web crawling lyrics websites for songs from bands identified as right-wing extremist by the German Federal Office for the Protection of the Constitution (Bundesamt für Verfassungsschutz)\footnote{See e.g., \citet[15]{vsb}, \href{https://www.bpb.de/themen/rechtsextremismus/dossier-rechtsextremismus/500809/rechtsrock/}{Bundeszentrale für politische Bildung}, \citet{dornbusch2002}}. The web scraping tool \texttt{Trafilatura} was used \cite{Barbaresi2021}.

This corpus contains a total of 99 songs. We selected 11 songs each from nine bands: \textit{Absurd, Landser, Stahlgewitter, Frontalkraft, Faustrecht, Nordfront, Die Lunikoff Verschwörung, Hassgesang}, and \textit{Sleipnir}. The temporal distribution of the songs is as follows: 22 songs from the 1990s, 57 from the 2000s, and 20 from the 2010s. The selection criteria for these bands were their activity during a similar period to \textit{Frei.Wild}, their official designation as right-wing extremist, and the online availability of their lyrics.

\subsection{The Reference Corpus for German Rock}
As a reference for general German rock music, we sampled from a larger corpus described by \citet{Schmidt2020}\footnote{Due to the license of the lyrics, that corpus is only available via e-mail request.}. This corpus contains 99 songs, with 11 songs each from nine well-known German rock bands: \textit{Annenmaykantereit, Die Ärzte, Großstadtgeflüster, Jennifer Rostock, OK KID, Rammstein, Sportfreunde Stiller, Tokio Hotel}, and \textit{Marius Müller-Westernhagen}. The bands were selected to match the temporal distribution of the Rechtsrock corpus, with 22 songs from the 1990s, 55 from the 2000s, and 22 from the 2010s.

\subsection{The \textit{Frei.Wild} Corpus}
The entire discography of \textit{Frei.Wild} is 303 songs as of December 2025. This is not counting covers, songs in English, instrumental songs, and songs that appeared on multiple albums. Of those, we sampled 25 from the 2000s and 22 from the 2010s to create a balanced corpus.\footnote{The reason that there are so few songs from the 2000s in the sample is that there were misguiding release dates on the official \textit{Frei.Wild} website. A later investigation, prompted by reviewers requesting classification results for all available \textit{Frei.Wild} songs, showed that these are not the original dates. \textit{Frei.Wild} had re-released most of their earlier albums under a new label in 2010.}

The rationale for including material from the 1990s in the two reference corpora, despite the fact that \textit{Frei.Wild}'s first official releases appeared in the early 2000s, is that this earlier period likely represents the formative years during which the band was influenced to create music. Since the aim of this study is to determine which genre \textit{Frei.Wild} most closely aligns with, we consider both the music that may have influenced them and the music that was released at the same time.

\section{Methods}
\label{sec:methods}
The aim of this study is to use computational methods to analyze the similarities between right-wing extremist rock music and the band \textit{Frei.Wild}.
\paragraph{Lexical analyses.} 
We begin our analysis at the word level. Following previous qualitative studies on Rechtsrock (see Section~\ref{sec:background}), we first examined the most frequent lemmata to obtain an overview of the dominant themes in the lyrics. To further examine the lexical findings, we applied a more targeted approach by analyzing concordances of the variations of `Heimat' used by the bands, and compared these findings to those reported by \citet{hofig2000}.

\paragraph{Classification.} From this focused investigation, that still required manual analysis of the findings, we then moved on to a broader approach to the question of \textit{Frei.Wild}'s affiliation. Both \citet{hartung2017} and \citet{dragos2023} achieved high performance for their classifiers with TF-IDF vectors, therefore, using a TF-IDF model of the lyrics, similarity scores were calculated between every band, and a binary classifier trained on the Rechtsrock and German Rock reference corpora. The classifier was then set to predict the category of the \textit{Frei.Wild} songs to discover how their songs are distributed. Due to the limitations of TF-IDF we then also used a sentence embedding model for German, to incorporate context and word meaning in the classification process.

Furthermore, for both classification experiments, the combinations of two genres (\textit{Frei.Wild} vs. Rechtsrock, and \textit{Frei.Wild} vs. German Rock) were tested to see which genre \textit{Frei.Wild} would be the hardest to separate from.

\paragraph{Temporal analysis.} With the skinhead background of the lead singer of \textit{Frei.Wild}, it is reasonable to assume that their earlier work might be more right-wing extremist and thereby skew the classification results. This is why we added a temporal dimension in this final analysis. The prediction results for each \textit{Frei.Wild} song from both classification analyses were plotted by year and then compared.

\paragraph{Preprocessing.} The lyrics were lemmatized and stripped of stop words and non-alphabetic tokens using the trained pipeline for German from \texttt{spaCy} [Version 3.7.1] \citep{spacy}.

\section{Lexical analyses}
\label{sec:lexical}
We began by compiling the lyrics from each corpus into three separate frequency lists and extracting the ten most frequent lemmata. Beyond the frequency patterns, we examined how the concept of `Heimat' (homeland) is used across genres, given its prominence in Rechtsrock and for the band \textit{Frei.Wild} \citep{hofig2000,Fischer2022,Hindrichs2014}. In Rechtsrock, `Heimat' is found to co-occur with words such as blood, honor, and pride. To trace such associations, we analyzed the most frequent concordances of `Heimat' as well as co-referential nouns across the three genres. The co-referential terms considered were `Vaterland' (fatherland), `Deutschland' (Germany), `Heimatland' (home country), and `Südtirol' (South Tyrol). The context window consisted of five words to the left and five words to the right of the target.

\paragraph{Results.}
The ten most frequent lemmata from the Rechtsrock corpus were: `volk', `deutsch', `stehen', `leben', `sehen', `land', `weg', `deutschland', `bleiben', `geben' (people/nation, German, stand, live, see, country, (a)way, Germany, stay, give), while in the German rock corpus they were: `hab', `leben', `liebe', `raus', `mal', `wissen', `sehen', `letzter', `schön', `sagen' (have, live, love, out, once, know, see, last, beautiful, say).
Lastly, the most frequent lemmata for \textit{Frei.Wild} are: `leben', `frei', `sehen', `bleiben', `stehen', `geben', `weg', `welt', `feind', `wild' (live, free, see, stay, stand, give, (a)way, world, enemy, wild).

Such frequency patterns provide a good overview of what the themes of the three corpora are. They all mention `(to) live' indicating the theme of their life experience, however in the Rechtsrock corpus this is accompanied by nationalistic words (nation, German, Germany): the concept of being a nation and staying this way (`bleiben') is in the foreground. The \textit{Frei.Wild} corpus also mentions `bleiben' and an enemy, which could be hinting at the resisting oppression narrative mentioned in \citet{Moeller2020}. The general German rock corpus mentions love, knowledge and beauty, creating more positive imagery and no specific narrative. We now turn to the concordance analyses, illustrated in Appendix \ref{sec:appendixCooc}.

Frequent collocates of `Vaterland' (fatherland) in Rechtsrock include `stehen', `blut', `schweiß', `träne' (stand, blood, sweat, tear). For `Deutschland' (Germany), the most common collocates are `schwarz', `rot', `gelb', `erheben', `Deutschland', `ruin', `volk' (black, red, yellow, raise, Germany, ruin, nation). In contrast, `Heimat' (homeland) co-occurs with `halten' and `land' (hold/keep, country/land), and overall appears less frequently than the other two terms. The collocational patterns suggest that `Heimat' is connected to possessiveness/protectiveness (hold, raise), to the country/the ground one is in/on (land, stand) as well as to violent/nationalistic terms (blood, nation, tears, ruin).

In the German rock corpus, `Heimat' only appears three times in the context of `Haus' (house), while `Deutschland' appears once; the remaining terms are absent. This indicates that the concepts found in Rechtsrock are absent in this genre.

In the \textit{Frei.Wild} corpus, `Heimat' occurs only twice. Instead, the band favors the compound `Heimatland' (homecountry) and the regional identifier `Südtirol' (South Tyrol), both of which do not appear in the other two corpora. These terms co-occur with words such as `sieger', `schlacht', `stolz' `brüdern entrissen', `fahne', `hölle schmorn', `feind' (winner, battle, pride, wrested [from its] brothers, flag, rot in hell (idiomatic), enemy). The parallels between these collocations and those found in the Rechtsrock corpus suggest a shared framing of homeland-related concepts.

\paragraph{Discussion.}
With the frequent lemmata, we find nationalistic themes with emphases on loyalty for the Rechtsrock genre, and the start of an enemy construction and resistance in the \textit{Frei.wild} corpus. This matches with the analysis of \citet{Moeller2020} (see Section~\ref{sec:background}), except for the use of violent imagery that one would expect for the Rechtsrock data. When looking at frequent lemmata from the separate bands on the Rechtsrock corpus, these do emerge in tokens like `tod', `blut', `schreie', `soldat', and `träne' (death, blood, screams, soldier, tear), nevertheless, what they all have in common are the nationalistic narratives also present in \textit{Frei.Wild} songs \citep{alt2020}. These were not as visible in the lemma analysis, however they emerged in the concordances.
Next to the associations mentioned in Section~\ref{sec:intro} for the Rechtsrock corpus, we found that \textit{Frei.Wild} and other Rechtsrock bands conceptualize their `Heimat' similarly. Looking at `Heimat' alone would not have provided these results, and it might seem justified to then conclude that there are actually fewer similarities because the two corpora do not share the same co-referential expressions. However, while it appeared to be relevant in previous qualitative studies, it was not the word specifically but the concept; while they sing about their `Heimat', the country/region that they are from, they do not frequently use that word to describe it. So it is not the case that Rechtsrock bands and \textit{Frei.Wild} talk about `Heimat' the same way, rather, they both view their respective homeland through a right-wing extremist lens that features violent imagery, nationalistic ideas, and an `us against them' narrative.

\section{Classification Experiments and Results}
\label{sec:classification}

Previous lexical analyses suggest that \textit{Frei.Wild} tends to be right-leaning, a finding consistently supported by critics and qualitative studies. Going beyond analyses that are restricted to specific words or concepts, we examine how this finding manifests at a broader level across bands and songs. While \textit{Frei.Wild} shares similarities with the Rechtsrock genre, open questions remain: To which bands in the corpus is it most similar, which of their songs appear most right-leaning, and how many such cases exist? To address this, we first computed inter-band similarity using a TF–IDF representation of the lyrics and cosine similarity. Building on this representation, we trained a random forest classifier and an SVM classifier on two reference corpora to predict whether a given song belongs to Rechtsrock or German Rock and applied this trained model to predict labels for the \textit{Frei.Wild} songs. Given that their performance did not differ significantly, we report only the results from the random forest classifier. Finally, going beyond isolated lexical features, we repeated the experiment with a sentence embedding representation to better capture word meaning in context.

All reported results come from a single run.

\subsection{Band Level Similarity: TF-IDF}
As discussed in Section~\ref{sec:lexical}, \textit{Frei.Wild} exhibits certain similarities to the Rechtsrock corpus. Since individual Rechtsrock bands may have different thematic foci, we conduct a pairwise comparison to identify where these similarities arise and whether they are limited to specific bands. To this end, we construct a TF–IDF representation for each band using the \texttt{Gensim} library [version 4.3.3] \citep{rehurek_lrec}, and compute a similarity matrix based on their cosine similarity values.

\begin{figure}[h]
  \includegraphics[width=\columnwidth]{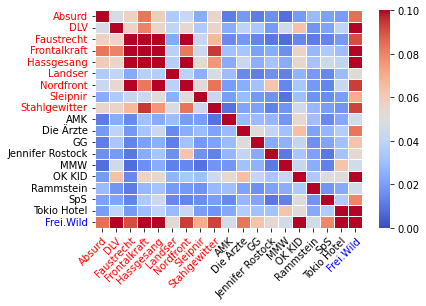}
  \caption{A heatmap showing the cosine similarities between bands from the Rechtsrock corpus (\textcolor{red}{red}), the Deutschrock corpus (black) and the band \textit{Frei.Wild} (\textcolor{blue}{blue}). Acronyms: `DLV' = \textit{Die Lunikoff-Verschwörung}; `AMK' = \textit{AnnenMayKantereit}; `GG' = \textit{Großstadtgeflüster}; `MMW' = \textit{Marius Müller-Westernhagen}; `SpS' = \textit{Sportfreunde Stiller}}
  \label{fig:heatmap_band}
\end{figure}

\paragraph{Results.} The similarity matrix is shown in Figure~\ref{fig:heatmap_band}.\footnote{All figures were created using the \texttt{seaborn} [Version 0.13.2] \citep{seaborn} and \texttt{matplotlib} [Version 3.8.2] \citep{matplotlib} libraries.}

\paragraph{Discussion.}

The matrix reveals comparatively higher internal similarity within the Rechtsrock corpus (a cluster is present at the top left of the matrix with the top nine bands, from \textit{Absurd} to \textit{Stahlgewitter}) than within the general rock corpus (a cluster is absent at the bottom right of the matrix with the middle nine bands, from \textit{AnnenMayKantereit} to \textit{Tokio Hotel}), which can be attributed to the narrower thematic focus of the former. The \textit{Frei.Wild} corpus (bottom band) stands out by exhibiting strong similarity to both corpora, with an arguably closer alignment to the Rechtsrock bands\footnote{The average similarity score between \textit{Frei.Wild} and all the Rechtsrock bands is 0.1, while for \textit{Frei.Wild} and the general German rock it is 0.07.}. This finding lends support to claims that \textit{Frei.Wild} is at least adjacent to right-wing extremist music.

\subsection{Classification: TF-IDF}\label{sec:classtfidf}
The findings so far suggest that \textit{Frei.Wild} positions between Rechtsrock and general German rock. To better establish which \textit{Frei.Wild} songs are the cause of the similarity to the Rechtsrock genre, a random forest classifier \citep{breiman2001} was trained using \texttt{scikit-learn} [Version 1.3.2] \citep{scikit-learn} on both reference corpora. The default parameters were kept in this analysis, setting \texttt{n\_estimators=100} (the number of trees) and \texttt{max\_features=`sqrt'} (the number of features considered for the split). %
The binary classifier was trained on TF-IDF representations of the Rechtsrock and general German rock data with a 75/25 train-test split and achieved an Accuracy score of 94\% and a ROC-AUC score of 95\%. No signs of overfitting were found by means of visualizing the classification probabilities on the training data (see Figure~\ref{fig:density_tfidf_training} in Appendix~\ref{sec:appendixTestpred}). Cross-validation (5-fold) resulted in a mean Accuracy of 82\% with a standard deviation of 0.06. Next to the results for the random forest classifier, we repeated the same classification task with 5-fold cross-validation with a SVM classifier: 82\% mean Accuracy with a standard deviation of 0.10, indicating a robustness of the methodology.
The TF-IDF transformed \textit{Frei.Wild} lyrics were then presented to this classifier as unlabeled data to be classified as either Rechtsrock or general German rock. We examine both the classification labels at a decision-threshold of 0.5 as well as the classification probabilities.

\begin{figure}[h]
  \includegraphics[width=\columnwidth]{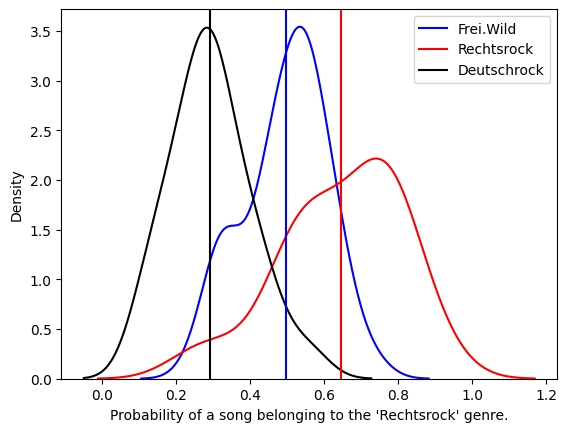}
  \caption{A density plot with means of the test data predictions from the TF-IDF classifier for Rechtsrock (\textcolor{red}{red}), Deutschrock (black) and the unlabeled data from \textit{Frei.Wild} (\textcolor{blue}{blue}).}
  \label{fig:density_tfidf}
\end{figure}

\paragraph{Results.} From the 47 \textit{Frei.Wild} songs, 51.06\% are classified as Rechtsrock (24 out of 47). 
While the classifier assigns roughly half of the data points to each class respectively, this alone does not reveal how confident the model is in its predictions. Figure~\ref{fig:density_tfidf} visualizes the classification probabilities and shows a unimodal distribution centered around 0.5 for \textit{Frei.Wild} data, 0.65 for Rechtsrock, and 0.3 for the general German rock data. This indicates that the model is generally uncertain about \textit{Frei.Wild} songs.

In addition, we trained the classifier on Rechtsrock vs \textit{Frei.Wild} (ROC-AUC: 89\%) and general German rock vs \textit{Frei.Wild} (ROC-AUC: 98\%). These classification scores suggest that the classifier can better separate \textit{Frei.Wild} from general German rock than from Rechtsrock.

For the (unbalanced) complete \textit{Frei.Wild} corpus, the results are similar: \textit{Frei.Wild} lyrics are classified as right-wing extremist music in 40.26\% of cases (122 of 303). A table of all songs and their right-wing-ness are provided in Appendix \ref{sec:appendixFull}.
\paragraph{Discussion.}
These results again suggest that \textit{Frei.Wild} is a true border case, with the classifier labeling half the data as Rechtsrock, and the unimodal distribution of the classification probabilities around 50\% suggesting high uncertainty in the classification. Nonetheless the \textit{Frei.Wild} mean is closer to the Rechtsrock mean than the general German rock mean. When we directly trained a classifier to separate \textit{Frei.Wild} from the two other classes, \textit{Frei.Wild} and Rechtsrock were harder to separate which indicates they are more similar.

\subsection{Classification: Sentence Embeddings}
While the TF-IDF representations offer relatively straightforward interpretability, they are inherently limited to capturing the frequency of isolated words. Since differences between the corpora may arise from variations in the contextual usage of the same words, we employ a sentence transformer model for the binary classification task to better incorporate semantic context. 
Each song was transformed into an embedding representation using the \href{https://huggingface.co/aari1995/German_Semantic_STS_V2}{\texttt{German\_Semantic\_STS\_V2}} model \citep{chibb2024german}, which was finetuned with the base model \texttt{gBERT-large} \citep{deepset}, and the \texttt{sentence-transformers} [Version 4.1.0] library \citep{sentence-transformers}.

The procedure is the same as in Section~\ref{sec:classtfidf}. A random forest classifier was trained on the embeddings of the two reference corpora with an Accuracy of 87.5\% and a ROC-AUC score of 97\%. No signs of overfitting were found by means of visualizing the classification probabilities on the training data (see Figure~\ref{fig:density_embedding_training} in Appendix~\ref{sec:appendixTestpred}). Cross-validation (5-fold) resulted in a mean Accuracy of 85\% with a standard deviation of 0.04. Next to the results for the random forest classifier, we repeated the same classification task with 5-fold cross-validation with an SVM classifier: 83\% mean Accuracy with a standard deviation of 0.08, indicating a robustness of the methodology.
The trained model predicted labels for the \textit{Frei.Wild} songs.

\begin{figure}[h]
  \includegraphics[width=\columnwidth]{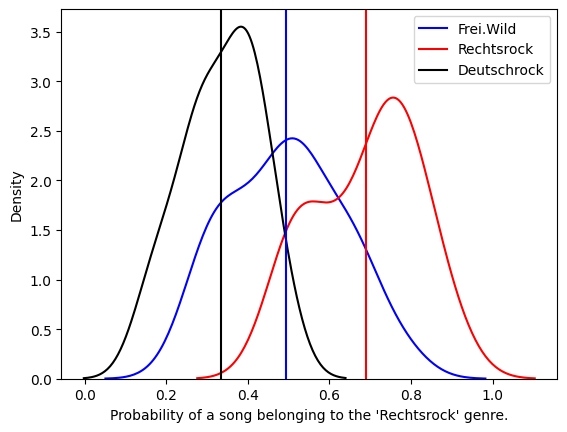}
  \caption{A density plot with means of the test data predictions from the sentence embedding classifier for Rechtsrock (\textcolor{red}{red}), Deutschrock (black) and the unlabeled data from \textit{Frei.Wild} (\textcolor{blue}{blue}).}
  \label{fig:density_embedding}
\end{figure}

\paragraph{Results.} \textit{Frei.Wild} lyrics are now classified as Rechtsrock in 59.57\% of cases. This shows that \textit{Frei.Wild} is right-leaning, as also visible in the density plot in Figure~\ref{fig:density_embedding}, showing a unimodal distribution centered around 0.5 for \textit{Frei.Wild} data, 0.69 for Rechtsrock, and 0.33 for general German rock. This indicates that the model is generally uncertain about \textit{Frei.Wild} songs.

In addition, we again trained the classifier on Rechtsrock vs \textit{Frei.Wild} (ROC-AUC: 77.5\%) and general German rock vs \textit{Frei.Wild} (ROC-AUC: 89\%). These classification scores suggest that the classifier can better separate general German rock and \textit{Frei.Wild}, than \textit{Frei.Wild} and Rechtsrock. 

For the (unbalanced) complete \textit{Frei.Wild} corpus, the results are similar: \textit{Frei.Wild} lyrics are classified as right-wing extremist music in 47.85\% of cases (145 of 303).

\paragraph{Discussion.} 
Figure~\ref{fig:density_embedding} shows that \textit{Frei.Wild} is still situated between the two reference corpora, similarly to the previous experiment. There is, however, a shift in the classification results: the proportion of \textit{Frei.Wild} songs labeled as Rechtsrock increases by nearly ten percentage points. In total, 14 songs received a different label, with nine changing from the general German rock category in the previous experiment to Rechtsrock. 

When we directly trained a classifier to separate \textit{Frei.Wild} from the other two classes, we again observed that \textit{Frei.Wild} and Rechtsrock were more difficult to separate, suggesting a higher degree of similarity. Taken together, these results indicate that when contextual information is included, \textit{Frei.Wild} tends to be associated more strongly with right-wing rock.

\section{Temporal Analysis}
The previous results show that the classifiers are uncertain with a tendency to right-wing extremist regarding \textit{Frei.Wild} songs. This could simply be the result of \textit{Frei.Wild} starting out more right-leaning, and after facing public backlash becoming more similar to mainstream German rock and less nationalistic. The density plots in Figures \ref{fig:density_tfidf} and \ref{fig:density_embedding} show that most songs are ambiguous but there is also a little bump towards Deutschrock. It would be fair to investigate if this is caused by the more recent songs because this then would mirror the band's own statements regarding their political affiliation. 
The classification scores for each \textit{Frei.Wild} song were gathered in a table, separated by model (TF-IDF and sentence embedding) and then visualized in a density plot by year.

\begin{figure}[h]
  \includegraphics[width=\columnwidth]{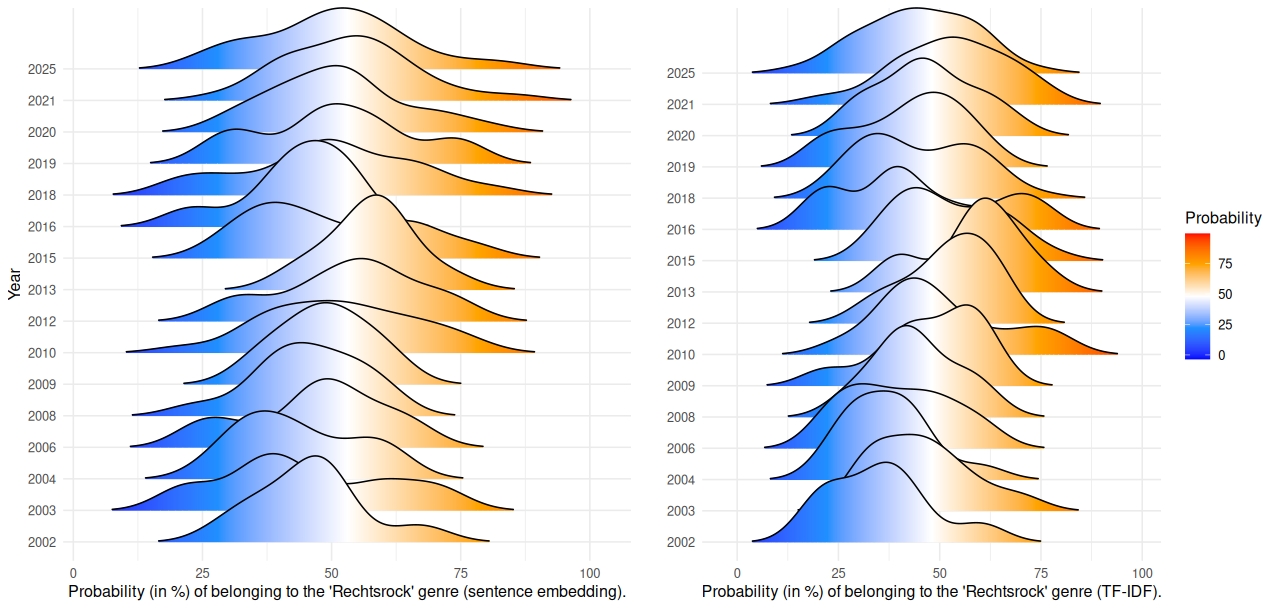}
  \caption{Two density plots for the classification scores of the \textit{Frei.Wild} songs per year, separated by model.}
  \label{fig:density_temporal}
\end{figure}

\paragraph{Results.} The plot is shown in Figure \ref{fig:density_temporal}. 

\paragraph{Discussion.} The temporal plots reveal that \textit{Frei.Wild} started out as less right-wing extremist, grew very extremist in the 2010s and has now settled to the right of the center, according to classification scores. This is unexpected as it contradicts \textit{Frei.Wild}'s self-portrayal and it also makes the strong association with right-wing rock more relevant.

\section{General Discussion}
\label{sec:discussion}
Looking back at the lexical analyses in Section~\ref{sec:lexical} and the classification experiments in Section~\ref{sec:classification}, we can confirm many of the intuitions that concern previous studies on Rechtsrock and the border case \textit{Frei.Wild}.
\textit{Frei.Wild} exhibits enemy/`us against them' constructions and an appeal to resistance tied to nationalistic narratives such as preserving certain values and the region/country that they call home, themes that they share with the Rechtsrock genre (cf. \citet{Moeller2020,alt2020,Fischer2022}).
Additionally, following the computational approaches by \citet{hartung2017} and \citet{dragos2023}, similarity measures and classification experiments revealed that there are not just a few \textit{Frei.Wild} songs that can be interpreted as Rechtsrock. Rather, there is a lot of internal similarity to different bands from the Rechtsrock genre and in a random sample of their lyrics, at least half of the songs would be classified as right-wing extremist. \textit{Frei.Wild} repeatedly distance themselves from those categorizations \citep{Hindrichs2014}, and we do find almost as much resemblance to the general German rock corpus as to Rechtsrock, however, any affinity to music that is judged as `youth-endangering' and unconstitutional should be seen as negative. \textit{Frei.Wild} moves in an ambiguous space with most of their songs. It is not the case that half is firmly German rock and the other half is extremist; mostly they are situated in the middle with a tendency towards Rechtsrock.
This quantitative approach therefore backs qualitative studies that do not interpret \textit{Frei.Wild} as politically neutral. The songs that were mentioned as potentially right-wing in these studies or judgments (cf. \citet{alt2020} or \citet[p. 20, `Federal Agency for Child- and Youth Media Protection']{bpjm}) were, for example, `Wahre werte', \textbf{`Land der Vollidioten'}, `Allein nach vorne', `Deutschrock ist Leidenschaft', \textbf{`Südtirol'}, \textbf{`Gutmenschen und Moralapostel'}, \textbf{`Schlagzeile groß Hirn zu klein'}, `Antiwillkommen', and \textbf{`Wir reiten in den Untergang'}, and songs from the first album `Eines Tages', especially the song `Rache muss sein'. Many of those criticized songs are not in the analyzed sample, but the ones that are (in \textbf{bold}) were also classified as Rechtsrock in both classification experiments, i.e., independent of input representation.
In our random sample, even without the inclusion of all songs that are deemed as right-wing by journalists or scientists, \textit{Frei.Wild} still has more in common with Rechtsrock than general German rock in this analysis. When the full discography is included, the scores are a bit lower, though still around half of the songs are classified as Rechtsrock. Further, the classification as right-wing extremist increases over time for \textit{Frei.Wild} (cf. Figure \ref{fig:density_temporal}), contradicting their claim of leaving their skinhead past behind. 

In an effort to increase generalizability of the results, a 95 song sample analysis of the similarly controversial band \textit{Böhse Onkelz} (Appendix \ref{sec:appendixBO}) revealed that their lyrics are classified as right-wing extremist in only 33.68\% of cases. They were unambiguously right-leaning in the 1980s and now argue to have changed \citep{Richter2006}; the classification experiments suggest that this is actually the case. The contrast to \textit{Frei.Wild} is especially apparent in the temporal analysis (cf. Figure \ref{fig:density_temporal_BO}), where \textit{Böhse Onkelz} song lyrics are for the most part classified as more general German rock since their comeback in 2013. 
Another contrast is the band \textit{Rammstein}. They had to publicly defend themselves against ``Nazi-accusations'' as well \citep{rs2011}, due to incorporating `NS aesthetics' (a.o. uniforms and excerpts from the 1936 summer Olympics propaganda) and disturbing themes, however, critics have argued that there is little indication of ideological commitment, as nationalist themes are largely absent from their music \citep{zizek2010} and they are also not indicated as a border case in our analysis. 
The statistical method of the Leave One Out approach described in Appendix~\ref{sec:appendixLOO}, which serves as a feature importance analysis, clears up how to distinguish between the categories Rechtsrock and general German rock for \textit{Frei.Wild}. The high impact words that drove classification are a.o. `frei' (free), `eu(e)r' (your, 2nd person plural), `lüge' (lie), `tod' (death) and `niemals' (never), underlining that it is not extreme imagery that is relevant for Rechtsrock but extremist narratives. In this context, the judgment of \textit{Frei.Wild} as a right-wing adjacent band seems even firmer.

\section{Conclusions and Future Work}
\label{sec:conclusion}
In this study, we presented lexical analyses and classification experiments aimed at determining whether the band \textit{Frei.Wild} should be categorized as politically right-leaning or as a part of the general German rock genre. Both resulted in settling \textit{Frei.Wild}'s position as a border case, as they share characteristics of Rechtsrock as well as German rock. Concerning their ambiguity, which is not to be judged as neutrality, they have usually slightly more similarities with Rechtsrock, such as the nationalistic `us against them' narratives, the higher similarity scores with the Rechtsrock bands, slightly more than 50\% of their songs labeled as right-wing extremist in the two classification experiments even with some of the more notorious songs missing in the sample.
The results of this study thereby confirm previous observations from qualitative studies concerned with Rechtsrock and \textit{Frei.Wild} \citep{Moeller2020,alt2020}, using methods commonly applied in computational studies on extremist data \citep{hartung2017,dragos2023}.
In the future, further analysis with an extended data set could provide clearer results, that is,
bigger reference corpora for Rechtsrock and general German rock, as well as possibly another reference corpus of other extremist music \citep{Moeller2020} to disentangle right-wing ideology specific aspects (e.g. white supremacy and antisemitism) from what constitutes extremism in general (e.g. inciting violence against an out-group \citep{berger2018}) \citep{hausler2021}. To support such extensions, we make our code and data available.

\section*{Limitations}
Several limitations impacted the study, first of all the size of the corpora, which was limited due to the (partially) government restricted public availability of right-wing extremist song lyrics. Restricted availability and only using bands labeled by authorities might limit the capabilities of the corpus to represent the full diversity of the genre. Other aspects of the three corpora can be better accounted for, such as subgenres/artistic styles within Rechtsrock and German rock. Both the song text and sound features have been suggested to work together to create a sense of collective identity and purpose in fascist music \citep{Machin01112012}. However, this study focused only on the lyrics but not the acoustic and visual features such as melody, as the availability of the actual sound files for Rechtsrock is even more restricted than the textual data. This narrower, single-case design focusing mostly on lyrics from a single German band, limits broader methodological validation, though there were few language-specific design choices (only the choice of the sentence embedding model and the spaCy model, as well as the choice of synonyms for `Heimat' in the concordance analysis) for the pipeline, making most of it transferable to other languages.
Lastly, our lexical analyses did not employ statistical measures commonly used in corpus linguistics, such as keyness or mutual information. This decision was motivated both by the relatively small size of the corpora and by the primary aim of the analyses, which was to provide a qualitative overview of the dominant themes. 
While alternative classification architectures and hyperparameter tuning could be explored, a random forest with default parameters already achieves high classification performance. This was confirmed in the classification experiments in which we trained both a random forest classifier and an SVM classifier, and their performance did not differ. Future work could investigate more advanced representation and classification methods. In particular, transformer-based models such as BERT or other pretrained language models could be used to learn richer contextual representations, either through fine-tuned supervised classifiers or via few-shot prompting with large language models acting as classifiers. These approaches may better capture semantic nuances; however, they come with increased computational cost, data requirements, and reduced interpretability compared to the current random forest approach. Given our emphasis on providing quantitative data on \textit{Frei.Wild}'s positioning rather than classification performance, we leave the exploration of such architectures to future work.

\section*{Ethical and Societal Implications}

At a time, where right-wing and nationalistic parties are on the rise globally, it is important to monitor influential agents in that sphere. Rechtsrock bands are part of an early recruitment step into the scene. Next to their popularity indicating how socially acceptable extremist and xenophobic ideas are becoming, analyzing their lyrics reveals current narratives. Those can be used to construct counter-narratives, which are argued to be more effective than debating right-wing arguments head-on.
Using NLP to analyze Rechtsrock lyrics at a large scale supports monitoring institutions in their efforts to mitigate the rightward shift we are currently experiencing by facilitating and accelerating the detection of right-wing extremist content in music.

This case study specifically aims to provide means to identify those tendencies even in borderline cases, like the band Frei.Wild in Germany, who are less explicit in their lyrics.

The song lyrics in the corpus are not licensed for distribution and most of the right-wing extremist lyrics are `indiziert' (indexed), meaning that they were indicated as `youth-endangering' and subject to a ban on distribution even among adults. The corpus is therefore available with only the metadata of the songs while the lyrics are not released as part of the dataset; they can be shared upon request for research purposes.

The involved university does not require IRB approval for this kind of study, which uses publicly available data without involving human participants. We do not see any other concrete risks concerning dual use of our research results. Of course, in the long run, any research results on AI methods could potentially be used in contexts of harmful and unsafe applications of AI. But this danger is rather low in our concrete case.

\section*{CRediT authorship contribution statement}

We follow the CRediT taxonomy\footnote{ \url{https://credit.niso.org/}}.
Conceptualization: [CS, KT]; Data curation: [CS]; Formal Analysis: [CS, KT]; Investigation: [CS, KT]; Methodology: [CS, KT]; Supervision: [KT]; Visualization: [CS]; and Writing – original draft: [CS, KT] and Writing – review \& editing: [CS, KT].

\bibliography{anthology,custom}

\begin{thebibliography}{38}
\providecommand{\natexlab}[1]{#1}

\bibitem[{Alt(2020)}]{alt2020}
Max Alt. 2020.
\newblock \href {https://doi.org/10.1007/978-3-658-29706-0_10} {{Die
  Nationalisierung der deutschsprachigen Popmusik. Neurechte Themen im
  Popdiskurs}}.
\newblock In Michael Ahlers, Lorenz Gr{\"u}newald-Schukalla, Anita J{\'o}ri,
  and Holger Schwetter, editors, \emph{Musik {\&} Empowerment}, pages 163--177.
  Springer Fachmedien Wiesbaden, Wiesbaden.

\bibitem[{Balzer(2019)}]{balzer2019}
Jens Balzer. 2019.
\newblock \href {https://d-nb.info/1172997950} {\emph{Pop und Populismus:
  {\"u}ber Verantwortung in der Musik}}.
\newblock Hamburg : Edition Körber.

\bibitem[{Barbaresi(2021)}]{Barbaresi2021}
Adrien Barbaresi. 2021.
\newblock \href {https://doi.org/10.18653/v1/2021.acl-demo.15} {Trafilatura:
  {A} web scraping library and command-line tool for text discovery and
  extraction}.
\newblock In \emph{Proceedings of the 59th Annual Meeting of the Association
  for Computational Linguistics and the 11th International Joint Conference on
  Natural Language Processing: System Demonstrations}, pages 122--131, Online.
  Association for Computational Linguistics.

\bibitem[{Berger(2018)}]{berger2018}
J.~M. Berger. 2018.
\newblock \href {https://doi.org/10.7551/mitpress/11688.001.0001}
  {\emph{Extremism}}.
\newblock The MIT Press.

\bibitem[{Breiman(2001)}]{breiman2001}
Leo Breiman. 2001.
\newblock \href {https://doi.org/10.1023/A:1010933404324} {Random forests}.
\newblock \emph{Machine Learning}, 45(1):5--32.

\bibitem[{{Bundesamt für
  Verfassungsschutz}(2020)}]{BfV2020RechtsextremistischeErlebniswelt}
{Bundesamt für Verfassungsschutz}. 2020.
\newblock \href
  {https://www.verfassungsschutz.de/SharedDocs/hintergruende/DE/rechtsextremismus/rechtsextremistische-erlebniswelt-musik-und-kampfsport.html}
  {{Rechtsextremistische Erlebniswelt: Musik und Kampfsport}}.
\newblock Access 2025.09.11.

\bibitem[{{Bundeszentrale f{\"u}r Kinder- und Jugendmedienschutz}(2015)}]{bpjm}
{Bundeszentrale f{\"u}r Kinder- und Jugendmedienschutz}. 2015.
\newblock \href
  {https://www.bzkj.de/resource/blob/175974/09490400ad9fde34ed3bd624bc11ba3c/2015-01-jahresrueckblick-2014-pdf-data.pdf}
  {\emph{BPJM aktuell: Jahresrückblick 2014}}.
\newblock Forum-Verl. Godesberg.

\bibitem[{Büchner(2018)}]{buechner2018weltbuergertum}
Timo Büchner. 2018.
\newblock \href {https://d-nb.info/1147397279} {\emph{``Weltbürgertum statt
  Vaterland'': Antisemitismus im RechtsRock}}.
\newblock edition assemblage, Münster.

\bibitem[{Büchner(2019)}]{buchner2019}
Timo Büchner. 2019.
\newblock \href {https://d-nb.info/119419379X} {\emph{Der Begriff" Heimat" in
  rechter Musik: Analysen--Hintergr{\"u}nde--Zusammenh{\"a}nge}}.
\newblock Wochenschau Verlag.

\bibitem[{Chan et~al.(2020)Chan, Schweter, and M{\"o}ller}]{deepset}
Branden Chan, Stefan Schweter, and Timo M{\"o}ller. 2020.
\newblock \href {https://doi.org/10.48550/arXiv.2010.10906} {German’s next
  language model}.
\newblock In \emph{International Conference on Computational Linguistics}.

\bibitem[{Chibb(2024)}]{chibb2024german}
Aaron Chibb. 2024.
\newblock \href {https://huggingface.co/aari1995/German_Semantic_STS_V2}
  {German\_semantic\_sts\_v2}.

\bibitem[{Dornbusch and Raabe(2002)}]{dornbusch2002}
Christian Dornbusch and Jan Raabe. 2002.
\newblock \href {https://d-nb.info/962257656} {\emph{RechtsRock :
  Bestandsaufnahme und Gegenstrategien / Christian Dornbusch, Jan Raabe
  (Hg.).}}, 1. aufl. edition.
\newblock RAT. Unrast, Hamburg.

\bibitem[{Dragos and Constable(2023)}]{dragos2023}
Valentina Dragos and Yol{\`e}ne Constable. 2023.
\newblock \href {https://doi.org/10.23919/FUSION52260.2023.10224162}
  {Comparison of classification techniques for extremism detection in {F}rench
  social media}.
\newblock In \emph{2023 26th International Conference on Information Fusion
  (FUSION)}, pages 1--6.

\bibitem[{Farin(2005)}]{Farin2005}
Klaus Farin. 2005.
\newblock \href {https://d-nb.info/1323651284/34#page=155} {{Reaktion{\"a}re
  Rebellen--Skinheads, Rechtsrock, B{\"o}hse Onkelz}}.
\newblock In \emph{Jugendkulturen}, pages 155--170. Mattes Verlag
  Pädagogischen Hochschule Heidelberg.

\bibitem[{Fischer(2022)}]{Fischer2022}
Michael Fischer. 2022.
\newblock \href {https://d-nb.info/125254202X/34#page=48} {{`Heimat ist kein
  Ort, Heimat ist ein Gefühl' (Herbert Grönemeyer) -- Konstruktion von Heimat
  in deutschsprachigen populären Songs des 21. Jahrhunderts}}.
\newblock In \emph{Naturschutz und Heimat -- Konzepte für die Zukunft
  entwickeln}, pages 47--65. Bonn.

\bibitem[{Hartung et~al.(2017)Hartung, Klinger, Schmidtke, and
  Vogel}]{hartung2017}
Matthias Hartung, Roman Klinger, Franziska Schmidtke, and Lars Vogel. 2017.
\newblock \href {https://doi.org/10.1007/978-3-319-59569-6_40} {Identifying
  right-wing extremism in {G}erman {T}witter profiles: A classification
  approach}.
\newblock In \emph{Natural Language Processing and Information Systems}, pages
  320--325, Cham. Springer International Publishing.

\bibitem[{Hindrichs(2014)}]{Hindrichs2014}
Thorsten Hindrichs. 2014.
\newblock \href {https://doi.org/10.22029/jlupub-2860} {Heimattreue {P}atrioten
  und das `{L}and der {V}ollidioten' - {Frei.Wild} und die neue
  {D}eutschrockszene}.
\newblock In \emph{Typisch deutsch}, pages 153--183. Bielefeld:
  transcript-Verlag.

\bibitem[{H{\"o}fig(2000)}]{hofig2000}
Eckhart H{\"o}fig. 2000.
\newblock \href {https://d-nb.info/95901974X} {\emph{Heimat in der Popmusik:
  Identit{\"a}t oder Kulisse in der deutschsprachigen Popmusikszene vor der
  Jahrtausendwende}}.
\newblock Triga.

\bibitem[{Honnibal et~al.(2020)Honnibal, Montani, Van~Landeghem, and
  Boyd}]{spacy}
Matthew Honnibal, Ines Montani, Sofie Van~Landeghem, and Adriane Boyd. 2020.
\newblock \href {https://doi.org/10.5281/zenodo.1212303} {{spaCy:
  Industrial-strength Natural Language Processing in Python}}.
\newblock \emph{Zenodo}.

\bibitem[{Hunter(2007)}]{matplotlib}
J.~D. Hunter. 2007.
\newblock \href {https://doi.org/10.1109/MCSE.2007.55} {Matplotlib: A 2d
  graphics environment}.
\newblock \emph{Computing in Science \& Engineering}, 9(3):90--95.

\bibitem[{Häusler(2021)}]{hausler2021}
Alexander Häusler. 2021.
\newblock \href {https://doi.org/10.46499/1760.2184} {Was ist rechts und was
  extrem?}
\newblock \emph{Politikum}, 7(4):4--9.

\bibitem[{Kuban(2012)}]{kuban2012}
Thomas Kuban. 2012.
\newblock \href {https://d-nb.info/1032943149} {\emph{Blut muss flie{\ss}en:
  Undercover unter Nazis}}.
\newblock Frankfurt am Main: Campus Verlag.

\bibitem[{Machin and Richardson(2012)}]{Machin01112012}
David Machin and John~E. Richardson. 2012.
\newblock \href {https://doi.org/10.1080/17405904.2012.713203} {Discourses of
  unity and purpose in the sounds of fascist music: a multimodal approach}.
\newblock \emph{Critical Discourse Studies}, 9(4):329--345.

\bibitem[{M{\"o}ller and Mischler(2020)}]{Moeller2020}
Veronika M{\"o}ller and Antonia Mischler. 2020.
\newblock \href {https://doi.org/10.1017/cri.2020.27} {The soundtrack of the
  extreme: Nasheeds and right-wing extremist music as a “gateway drug” into
  the radical scene?}
\newblock \emph{International Annals of Criminology}, 58(2):291--334.

\bibitem[{Naumann(2009)}]{naumann2009}
Thomas Naumann. 2009.
\newblock \href {http://www.ciando.com/ebook/bid-277548} {\emph{Rechtsrock im
  Wandel. Eine Textanalyse von Rechtsrock-Bands.}}
\newblock Diplomica Verl., Hamburg.

\bibitem[{Neumann(2013)}]{stz2013}
Olaf Neumann. 2013.
\newblock \href
  {https://www.stuttgarter-zeitung.de/inhalt.interview-ueber-musikszene-freiwild-machen-eindeutig-rechtsrock.5d1ce6a2-4c57-41c8-8b9f-8a688c4155d8.html}
  {{Freiwild machen eindeutig Rechtsrock}}.
\newblock Accessed: 2025-09-22.

\bibitem[{Pedregosa et~al.(2011)Pedregosa, Varoquaux, Gramfort, Michel,
  Thirion, Grisel, Blondel, Prettenhofer, Weiss, Dubourg, Vanderplas, Passos,
  Cournapeau, Brucher, Perrot, and Duchesnay}]{scikit-learn}
F.~Pedregosa, G.~Varoquaux, A.~Gramfort, V.~Michel, B.~Thirion, O.~Grisel,
  M.~Blondel, P.~Prettenhofer, R.~Weiss, V.~Dubourg, J.~Vanderplas, A.~Passos,
  D.~Cournapeau, M.~Brucher, M.~Perrot, and E.~Duchesnay. 2011.
\newblock \href {https://doi.org/10.5555/1953048.2078195} {Scikit-learn:
  Machine learning in {P}ython}.
\newblock \emph{Journal of Machine Learning Research}, 12:2825--2830.

\bibitem[{{\v R}eh{\r u}{\v r}ek and Sojka(2010)}]{rehurek_lrec}
Radim {\v R}eh{\r u}{\v r}ek and Petr Sojka. 2010.
\newblock {Software Framework for Topic Modelling with Large Corpora}.
\newblock In \emph{{Proceedings of the LREC 2010 Workshop on New Challenges for
  NLP Frameworks}}, pages 45--50, Valletta, Malta. ELRA.
\newblock \url{http://is.muni.cz/publication/884893/en}.

\bibitem[{Reimers and Gurevych(2019)}]{sentence-transformers}
Nils Reimers and Iryna Gurevych. 2019.
\newblock \href {https://doi.org/10.48550/arXiv.1908.10084} {{Sentence-BERT:
  Sentence Embeddings using Siamese BERT-Networks}}.
\newblock In \emph{Proceedings of the 2019 Conference on Empirical Methods in
  Natural Language Processing}. Association for Computational Linguistics.

\bibitem[{Richter(2006)}]{Richter2006}
Stephan Richter. 2006.
\newblock \href
  {https://edoc.vifapol.de/opus/volltexte/2009/1249/pdf/band_27.pdf#page=110}
  {{„Gehasst--verdammt--verg{\"o}ttert“ -- Das Ph{\"a}nomen der ehemaligen
  Skinhead-Kultband „B{\"o}hse Onkelz“ und ihre Bez{\"u}ge zum
  Rechtsextremismus}}.
\newblock In \emph{Rechtsextremismus als Gesellschaftsphänomen --
  Jugendhintergrund und Psychologie}, pages 109--189. Fachhochschule des Bundes
  für öffentliche Verwaltung: Fachbereich Öffentliche Sicherheit.

\bibitem[{{Rolling Stone Germany}(2011)}]{rs2011}
{Rolling Stone Germany}. 2011.
\newblock \href
  {https://www.rollingstone.de/rammstein-exklusives-interview-mit-till-lindemann-und-flake-lorenz-343190/}
  {{Rammstein: Exklusives Interview mit Till Lindemann und Flake}}.
\newblock Accessed: 2025-10-02.

\bibitem[{Schmidt et~al.(2020)Schmidt, Bauer, Habler, Heuberger, Pilsl, and
  Wolff}]{Schmidt2020}
Thomas Schmidt, Marlene Bauer, Florian Habler, Hannes Heuberger, Florian Pilsl,
  and Christian Wolff. 2020.
\newblock \href {https://doi.org/10.5281/zenodo.4621928} {{Der Einsatz von
  Distant Reading auf einem Korpus deutschsprachiger Songtexte}}.
\newblock In Christof Sch{\"o}ch, editor, \emph{DHd 2020: Spielr{\"a}ume;
  Digital Humanities zwischen Modellierung und Interpretation.
  Konferenzabstracts; Universit{\"a}t Paderborn, 02. bis 06. M{\"a}rz 2020},
  pages 296--300. Zenodo, Paderborn, Germany.

\bibitem[{Stede and Memminger(2025)}]{stede2025}
Manfred Stede and Ronja Memminger. 2025.
\newblock \href {https://doi.org/10.18653/v1/2025.nlp4pi-1.14} {{A}f{D}-{CCC}:
  Analyzing the climate change discourse of a {G}erman right-wing political
  party}.
\newblock In \emph{Proceedings of the Fourth Workshop on NLP for Positive
  Impact (NLP4PI)}, pages 163--174, Vienna, Austria. Association for
  Computational Linguistics.

\bibitem[{{Verfassungsschutz Baden-Württemberg}(2024)}]{vsb2024}
{Verfassungsschutz Baden-Württemberg}. 2024.
\newblock \href
  {https://www.verfassungsschutz-bw.de/site/verfassungsschutz/get/documents_E716343085/IV.Dachmandant/LfV_Datenquelle_neu/Publikationen/Jahresberichte/Verfassungsschutzbericht%20Baden-W%C3%BCrttemberg%202024.pdf}
  {\emph{Verfassungsschutzbericht}}.
\newblock Ministerium des Innern, für Digitalisierung und Kommunen.

\bibitem[{{Verfassungsschutz Rheinland-Pfalz}(1996)}]{vsb}
{Verfassungsschutz Rheinland-Pfalz}. 1996.
\newblock \href
  {https://verfassungsschutzberichte.de/pdfs/vsbericht-rp-1996.pdf}
  {\emph{Tätigkeitsbericht 1996 des rheinland-pfälzischen
  Verfassungsschutzes}}.
\newblock Ministerium des Innern und für Sport, Mainz.

\bibitem[{Waskom(2021)}]{seaborn}
Michael~L. Waskom. 2021.
\newblock \href {https://doi.org/10.21105/joss.03021} {seaborn: statistical
  data visualization}.
\newblock \emph{Journal of Open Source Software}, 6(60):3021.

\bibitem[{Xie et~al.(2024)Xie, Ahia, Tsvetkov, and Anastasopoulos}]{xie2024}
Roy Xie, Orevaoghene Ahia, Yulia Tsvetkov, and Antonios Anastasopoulos. 2024.
\newblock \href {https://arxiv.org/abs/2402.17914} {Extracting lexical features
  from dialects via interpretable dialect classifiers}.
\newblock \emph{Preprint}, arXiv:2402.17914.

\bibitem[{{\v{Z}}i{\v{z}}ek(2010)}]{zizek2010}
Slavoj {\v{Z}}i{\v{z}}ek. 2010.
\newblock \href {https://doi.org/10.1057/sub.2009.34} {Some concluding notes on
  violence, ideology and communist culture}.
\newblock \emph{Subjectivity}, 3(1):101--116.

\end{thebibliography}

\appendix

\section{Lexical features with leave-one-out analyses}
\label{sec:appendixLOO}

The Leave One Out approach \citep{xie2024} determines words that are relevant for classification by checking if prediction accuracy decreases when a word is taken out of the sentence.
Words which were picked as relevant for Rechtsrock classification mirror previous qualitative analyses. Among the top features were `deutsch', `land', `kämpfen', `einst', `feind', `niemals', `frei', `krieg', `sieg', `volk', `vaterland', `blut', `tod', `lüge', and `eur' (German, country, fight, once, enemy, never, free, war, victory, nation, fatherland, blood, death, lie, and your (2nd person plural)), which are also featured in the most frequent lemmata of the Rechtsrock corpus and general extremist talking points (an in- and out-group concept with reference to violence against the out-group \citep{berger2018}).

As discussed in Section~\ref{sec:lexical}, the word `Heimat' (homeland) is not featured, however, the word `Vaterland' (fatherland) is, supporting our finding that this is the expression right-wing extremist rock music employs to refer to the concept of homeland that carries their ideological connotations.

\section{Example: Concordance Analysis}
\label{sec:appendixCooc}

An example from the \textit{Frei.Wild} corpus would be a list of concordances of the word `Südtirol' (same extension as `Heimat' for that band). The first element of the list is the left context, the second is the right context (five words each):
\texttt{[[`schlagzeug', `gitarre', `erklingen', `liedche', `singen'], [`tragen', `fahne', `schön', `land', `welt']]} \\
\texttt{[[ `stolz', `sohn', `heimatland', `geben', `niemehr'],  [`brüdern', `entrissen', `hinaus', `wissen', `südtirol']]}\\
\texttt{[[ `südtirol', `brüdern', `entrissen', `hinaus', `wissen'],  [`verlorn', `hölle', `feind', `schmorn', `heiß']]}\\
\texttt{[[ `luft', `hinaus', `heimatland', `geben', `niemals'],  [`tragen', `fahne', `schön', `land', `welt']]}\\
\texttt{[[ `hand', `schönerer', `ewigkeit', `erde', `bereit'],  [`heimatland', `herzstück', `welt', `liegen', `hand']]}\\
\texttt{[[ `berg', `geburtsort', `vieler', `held', `geben'],  [`leben', `verlierer', `schnell', `hemd', `schlimm']]}\\

For the Rechtsrock corpus, there were more concordances for `Deutschland' and therefore a counter provided this frequency list:
\texttt{`schwarz': 14, `rote': 12, `gelb': 10, `erheben': 9, `deutschland': 8, `ruin': 8, `volk': 7, `stehen': 7, `leben': 6, `nacht': 5, `stolz': 5}

\section{Classification probabilities of the training data}
\label{sec:appendixTestpred}

\begin{figure}[h]
  \includegraphics[width=\columnwidth]{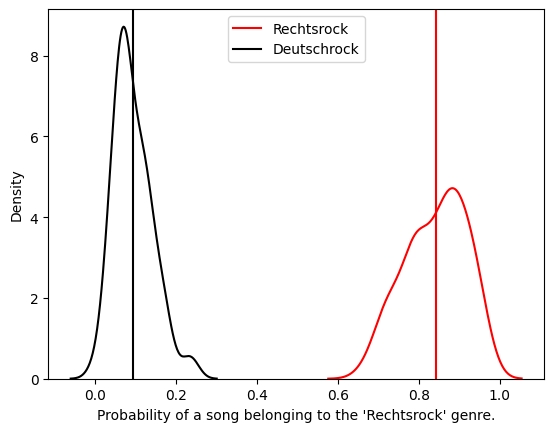}
  \caption{A density plot with means of the training data predictions from the TF-IDF classifier for Rechtsrock (\textcolor{red}{red}) and Deutschrock (black).}
  \label{fig:density_tfidf_training}
\end{figure}

\begin{figure}[h]
  \includegraphics[width=\columnwidth]{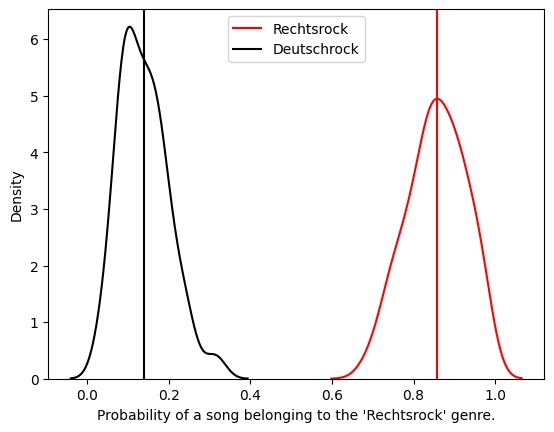}
  \caption{A density plot with means of the training data predictions from the sentence embedding classifier for Rechtsrock (\textcolor{red}{red}) and Deutschrock (black).}
  \label{fig:density_embedding_training}
\end{figure}

No signs of overfitting were found by means of visualizing the classification probabilities on the training data, as can be seen in the density plots in Figures \ref{fig:density_tfidf_training} and \ref{fig:density_embedding_training}.

\section{\textit{Böhse Onkelz} sample analysis}
\label{sec:appendixBO}

\begin{figure}[h]
  \includegraphics[width=\columnwidth]{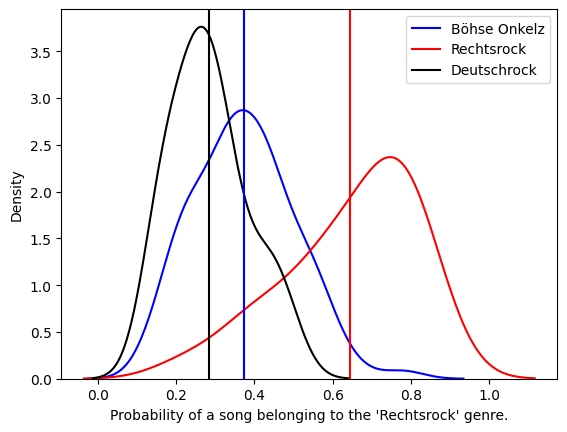}
  \caption{A density plot with means of the test data predictions from the TF-IDF classifier for Rechtsrock (red), Deutschrock (black) and the unlabeled data from \textit{Böhse Onkelz} (blue).}
  \label{fig:density_tfidf_training_BO}
\end{figure}

\begin{figure}[h!]
  \includegraphics[width=\columnwidth]{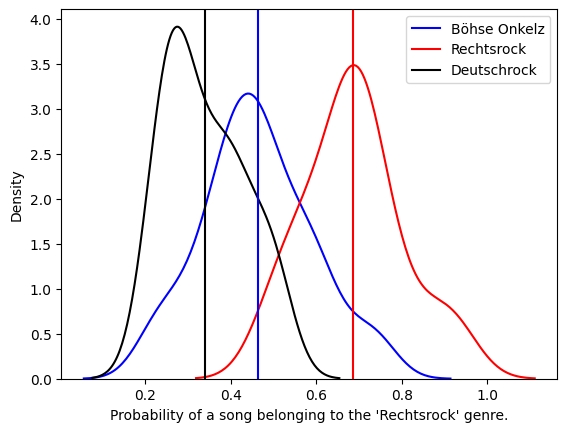}
  \caption{A density plot with means of the test data predictions from the sentence embedding classifier for Rechtsrock (red), Deutschrock (black) and the unlabeled data from \textit{Böhse Onkelz} (blue).}
  \label{fig:density_embedding_training_BO}
\end{figure}

The band \textit{Böhse Onkelz} is a border case for right-wing extremism categorization similar to \textit{Frei.Wild}. They are a skinhead band from 1980 that inspired a lot of Rechtsrock bands. They retired in 2005 and are back since 2013, while also claiming to have left their extremist past behind them.
The sentence embedding classifier, trained on the reference corpora (Rechtsrock and general German rock) with an Accuracy of 94\% and a ROC-AUC score of 99\%, classified their lyrics as right-wing extremist music in around 30\% of cases. They are less right-leaning than \textit{Frei.wild} in our analysis, which is also visible in the density plots in Figure \ref{fig:density_tfidf_training_BO} and \ref{fig:density_embedding_training_BO}.

\begin{figure}[h]
  \includegraphics[width=\columnwidth]{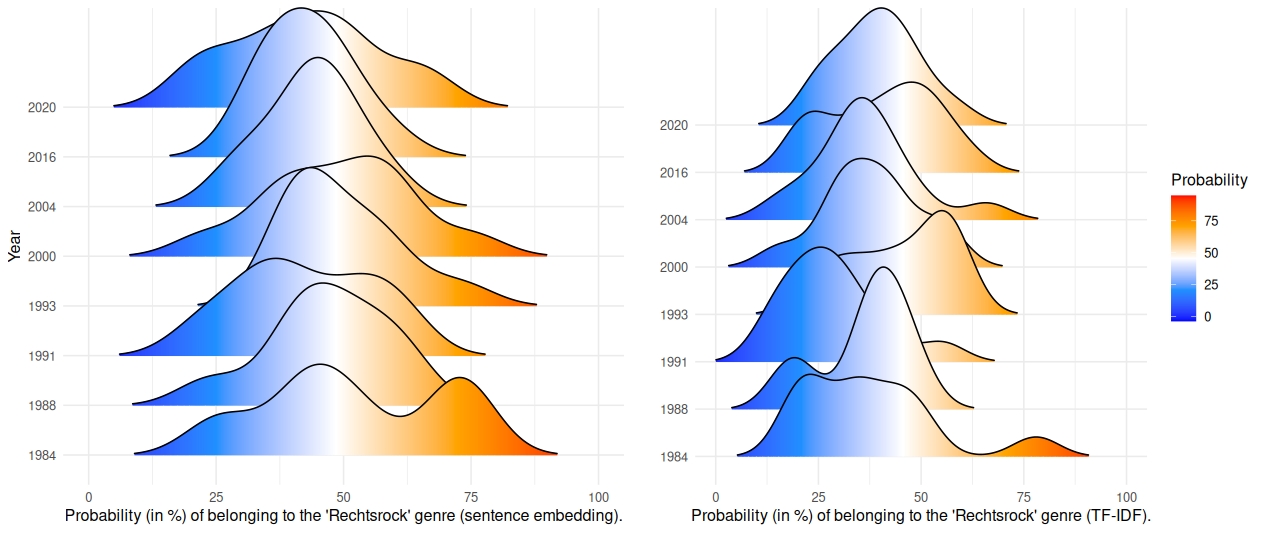}
  \caption{Two density plots for the classification scores of the \textit{Böhse Onkelz} songs per year, separated by model.}
  \label{fig:density_temporal_BO}
\end{figure}

The temporal analysis in Figure \ref{fig:density_temporal_BO} indicates that they indeed started out as a right-wing extremist rock band and moved to more mainstream and less nationalistic/xenophobic themes over time. This stands in direct contrast to the development in the temporal analysis of \textit{Frei.Wild} lyrics (cf. Figure \ref{fig:density_temporal}), which seem to include an increasing percentage of right-wing content.

\section{Song names and their English translations in order of appearance}
\label{sec:appendixsong}

\begin{table}[h!]
\small
\centering
\rotatebox{90}{
\begin{tabular}{ll}
\hline
\textbf{German Title} & \textbf{English Translation} \\
\hline
\textit{Sturmführer} & storm-leader (lowest officer rank in Nazi Germany) \\
\textit{Bis hier her} & this far (as in “so far and no further”) \\
\textit{Wahre Werte} & true values \\
\textit{Südtirol} & South Tyrol \\
\textit{Antiwillkommen} & anti-welcome \\
\textit{Gutmenschen und Moralapostel} & do-gooder and moralizer \\
\textit{Schlagzeile zu groß, Hirn zu klein} & headline too big, brain too small \\
\textit{Land der Vollidioten} & land of complete fools \\
\textit{Allein nach vorne} & alone forwards \\
\textit{Deutschrock ist Leidenschaft} & German rock is passion \\
\textit{Wir reiten in den Untergang} & we ride into doom/demise \\
\textit{Rache muss sein} & revenge is a must \\
\hline
\end{tabular}}
\caption{German song titles and their English translations.}
\label{tab:song_titles}
\end{table}

\onecolumn

\section{Full \textit{Frei.Wild} discography classification scores}
\label{sec:appendixFull}

\begin{table*}[h]
    \centering
    \begin{tabular}{|l|l|l|}
    \hline
        \textbf{\textit{Frei.Wild} song title} & \textbf{TF-IDF} & \textbf{sentence embedding} \\ \hline
        15 Jahre & 67 & 44 \\ \hline
        20 Jahre Seite an Seite & 47 & 57 \\ \hline
        AIDS & 42 & 35 \\ \hline
        Akzeptierter Faschist & 53 & 63 \\ \hline
        Alarm im Proberaum & 56 & 36 \\ \hline
        Alkohol & 44 & 44 \\ \hline
        Alle Menschen sind gleich & 62 & 51 \\ \hline
        Allein nach vorne & 81 & 78 \\ \hline
        Allein, Ohne Dich, Bei Mir & 41 & 50 \\ \hline
        Alles alles was mir fehlt & 31 & 54 \\ \hline
        Alles auf Offbeat & 41 & 54 \\ \hline
        Alles in allem  & 29 & 44 \\ \hline
        Alles ist weg & 15 & 33 \\ \hline
        Alles um uns ist still & 48 & 38 \\ \hline
        Altes neues Leben & 37 & 62 \\ \hline
        Antiwillkommen & 51 & 53 \\ \hline
        Arschloch, lass mich in Ruh! & 41 & 53 \\ \hline
        Arschtritt & 62 & 51 \\ \hline
        Attacke ins Glück & 45 & 64 \\ \hline
        Auf ein nie wieder Wiedersehen & 45 & 63 \\ \hline
        Auf einen Neuanfang & 37 & 54 \\ \hline
        Auf zum Schwur & 58 & 67 \\ \hline
        Auge um Auge, Zahn um Zahn & 35 & 44 \\ \hline
        Aus dem Film unserer Geschichte & 34 & 65 \\ \hline
        Aus Traum wird Wirklichkeit & 63 & 28 \\ \hline
        Außenseiterband & 62 & 51 \\ \hline
        B.O.U.L & 28 & 41 \\ \hline
        Betteln vs. Batteln & 54 & 44 \\ \hline
        Bilder und Narben, meine Memoiren & 42 & 55 \\ \hline
        Bin kein Held, aber Mensch bin ich & 23 & 42 \\ \hline
        Blinde Völker wie Armeen & 56 & 58 \\ \hline
        Böse und gemein & 31 & 49 \\ \hline
        Brixen & 45 & 45 \\ \hline
        Brüderlein zum Wohl & 25 & 39 \\ \hline
        Ciao Bella, Ciao & 44 & 50 \\ \hline
    \end{tabular}
    \caption{Rechtsrock probability (in \%) assigned to the individual \textit{Frei.Wild} songs by the classifiers. If the value was >50, they were classified as Rechtsrock.}
\end{table*}

\begin{table*}
\ContinuedFloat 
    \centering
    \begin{tabular}{|l|l|l|}
    \hline
        \textbf{\textit{Frei.Wild} song title} & \textbf{TF-IDF} & \textbf{sentence embedding} \\ \hline
        Corona Weltuntergang & 49 & 52 \\ \hline
        Corona Weltuntergang V2 & 61 & 56 \\ \hline
        Daheim & 22 & 47 \\ \hline
        Das gewisse Nichts & 36 & 45 \\ \hline
        Das Land der Vollidioten & 50 & 52 \\ \hline
        Dein zweites Leben & 47 & 21 \\ \hline
        Der aufrechte Weg & 49 & 25 \\ \hline
        Der Gast in deinem Geist & 32 & 44 \\ \hline
        Der Horizont weist uns die Richtung & 40 & 21 \\ \hline
        Der König ist tot, es lebe der König & 43 & 41 \\ \hline
        Der Staat vergibt dein Gewissen nicht & 57 & 72 \\ \hline
        Der Teufel trägt Geweih & 43 & 60 \\ \hline
        Der Tod, der holt uns alle & 57 & 49 \\ \hline
        Des Schicksals Schmied & 55 & 40 \\ \hline
        Deutschrock ist Leidenschaft & 52 & 55 \\ \hline
        Die Band, die Wahrheit bringt & 62 & 59 \\ \hline
        Die Gedanken sie sind frei & 64 & 50 \\ \hline
        Die Geister, die ich nicht rief & 50 & 58 \\ \hline
        Die Hölle schenkte uns das Licht & 58 & 63 \\ \hline
        Die Lösung vom Problem & 57 & 40 \\ \hline
        Die nur nach fremden Sünden fischen & 44 & 55 \\ \hline
        Die Welt brennt & 50 & 52 \\ \hline
        Die Zeit vergeht & 46 & 43 \\ \hline
        Diese Nacht will nicht meine Nacht sein & 22 & 35 \\ \hline
        Diesen Schuh musst du dir nicht anziehen & 59 & 39 \\ \hline
        Du bist ein Idiot & 28 & 63 \\ \hline
        Du bist Sie (Die Einzige für mich) & 42 & 34 \\ \hline
        Du kriegst nicht eine Sekunde zurück & 61 & 50 \\ \hline
        Dunkel, Hell, Schwarz und Grau & 54 & 51 \\ \hline
        Ebbe und Flut & 39 & 56 \\ \hline
        Echo, Platin und Gold & 52 & 61 \\ \hline
        Eine Freundschaft, eine Liebe, eine Familie & 53 & 73 \\ \hline
        Eines Tages & 21 & 36 \\ \hline
        Engel über dem Himmel & 32 & 43 \\ \hline
        Engel und Bengel & 32 & 35 \\ \hline
        Es braucht nicht viel um Glücklich zu sein & 69 & 68 \\ \hline
        Es geht hier um mein Leben & 29 & 41 \\ \hline
        Es gibt nicht nur einen Weg & 41 & 57 \\ \hline
        Es ist vorbei, es ist Geschichte & 40 & 49 \\ \hline
        Es ist Wahnsinn, es ist Liebe & 58 & 53 \\ \hline
        Europa & 43 & 48 \\ \hline
        Feinde deiner Feinde & 56 & 69 \\ \hline
        Feste fallen, wie sie fallen, aber landen tun sie hart & 49 & 36 \\ \hline
        Feuchte Augen & 35 & 50 \\ \hline
        Feuer, Erde, Wasser, Luft & 61 & 49 \\ \hline
        Fick dich und verpiss dich & 35 & 49 \\ \hline
    \end{tabular}
    \caption{Rechtsrock probability (in \%) assigned to the individual \textit{Frei.Wild} songs by the classifiers. If the value was >50, they were classified as Rechtsrock. (continued)}
\end{table*}

\begin{table*}
\ContinuedFloat 
    \centering
    \begin{tabular}{|l|l|l|}
    \hline
        \textbf{\textit{Frei.Wild} song title} & \textbf{TF-IDF} & \textbf{sentence embedding} \\ \hline
        1860 & 31 & 58 \\ \hline
        Frei.Wild & 35 & 72 \\ \hline
        Frei.Wild's Ländereien & 57 & 66 \\ \hline
        Freiheit & 60 & 48 \\ \hline
        Freiheit, Freundschaft, Brüderlichkeit & 74 & 63 \\ \hline
        Freundschaft & 45 & 38 \\ \hline
        Für Glaube, für Liebe, für Hoffnung & 40 & 32 \\ \hline
        Für immer Anker und Flügel & 75 & 72 \\ \hline
        Für immer und ewig unendlich & 30 & 50 \\ \hline
        Für immer untrennbar & 41 & 49 \\ \hline
        Für meine Freiheit & 45 & 61 \\ \hline
        Fürchte dich nicht & 37 & 49 \\ \hline
        Geartete Künste hatten wir schon & 75 & 72 \\ \hline
        Gebt mir die Pappe wieder & 36 & 35 \\ \hline
        Geile Heimat & 68 & 70 \\ \hline
        Gipfelstürmer & 50 & 51 \\ \hline
        Gladiator und Draufgänger & 67 & 73 \\ \hline
        Gleld & 33 & 41 \\ \hline
        Glückes Schmiedes Dieb & 42 & 75 \\ \hline
        Gott und wir selbst & 62 & 70 \\ \hline
        Gratissäufer & 23 & 33 \\ \hline
        Gutmensch ärgere dich nicht & 34 & 40 \\ \hline
        Gutmenschen und Moralapostel & 64 & 55 \\ \hline
        Hab keine Angst & 45 & 41 \\ \hline
        Halbstark, laut und jung & 63 & 55 \\ \hline
        Halt deine Schnauze & 44 & 59 \\ \hline
        Hart am Wind & 74 & 54 \\ \hline
        Harte Zeiten = Hartes Leben & 22 & 47 \\ \hline
        Heile mich, heile dich & 36 & 38 \\ \hline
        Heimat & 60 & 59 \\ \hline
        Heimat im Herzen und Neuland im Blut & 66 & 56 \\ \hline
        Heiße Kälte & 60 & 49 \\ \hline
        Herz schlägt Herz & 41 & 56 \\ \hline
        Hey, ich lebe noch & 29 & 32 \\ \hline
        Hier geboren und doch verloren & 64 & 51 \\ \hline
        Hier rein da raus Freigeist & 28 & 56 \\ \hline
        Hoch hinaus & 50 & 66 \\ \hline
        Hör, was das Herz dir sagt & 59 & 65 \\ \hline
        Ich & 34 & 62 \\ \hline
        Ich baue mir ein Denkmal & 68 & 70 \\ \hline
        Ich bin bereit & 35 & 34 \\ \hline
        Ich bin neu, ich fange an & 57 & 61 \\ \hline
        Ich bin nicht heilig & 31 & 59 \\ \hline
        Ich bleib daheim & 20 & 47 \\ \hline
        Ich denk an euch zurück & 42 & 36 \\ \hline
        Ich gebe euch allen nur n' Fuck & 26 & 30 \\ \hline
    \end{tabular}
    \caption{Rechtsrock probability (in \%) assigned to the individual \textit{Frei.Wild} songs by the classifiers. If the value was >50, they were classified as Rechtsrock. (continued)}
\end{table*}

\begin{table*}
\ContinuedFloat 
    \centering
    \begin{tabular}{|l|l|l|}
    \hline
        \textbf{\textit{Frei.Wild} song title} & \textbf{TF-IDF} & \textbf{sentence embedding} \\ \hline
        Ich helf dir auf & 58 & 42 \\ \hline
        Ich kann auf die Fresse geben & 58 & 59 \\ \hline
        Ich lache über dich & 37 & 45 \\ \hline
        Ich und mein Scheiss & 59 & 45 \\ \hline
        Ich weiß wer ich war & 57 & 53 \\ \hline
        Ich werde mich jetzt nicht ändern & 41 & 51 \\ \hline
        Ich will dich irgendwann verlieren & 36 & 42 \\ \hline
        Ich will leben & 55 & 60 \\ \hline
        Im Auftrag der Welt & 58 & 80 \\ \hline
        Im Zweifel nicht für Dich & 40 & 52 \\ \hline
        Immer höher hinaus & 50 & 53 \\ \hline
        Immer nur nach vorne & 28 & 40 \\ \hline
        Immer unter Feuer & 42 & 68 \\ \hline
        Imola & 52 & 60 \\ \hline
        In 8 Minuten um die Welt & 25 & 20 \\ \hline
        In der Mitte stehe ich & 62 & 66 \\ \hline
        Irgendwann & 40 & 44 \\ \hline
        Irgendwer steht dir zur Seite & 27 & 24 \\ \hline
        It's a good day for a good day & 46 & 49 \\ \hline
        Jedes Glück braucht auch sein Leid & 41 & 52 \\ \hline
        Jenseits von Milliarden & 48 & 56 \\ \hline
        Joanna an der Bar & 27 & 39 \\ \hline
        Junge mach weiter & 36 & 41 \\ \hline
        Kämpfer werden bis zum Ende gehen & 36 & 54 \\ \hline
        Kein Zoll zurück & 52 & 65 \\ \hline
        Keine Angst vor Liebe & 53 & 44 \\ \hline
        Keine Lüge reicht je bis zur Wahrheit & 48 & 48 \\ \hline
        Kick ass vs. Arschtritt & 61 & 45 \\ \hline
        Krieger des Lichts & 51 & 57 \\ \hline
        Land der Vollidioten & 45 & 49 \\ \hline
        Lass dein Leben niemals los & 62 & 46 \\ \hline
        Lass dich gehen & 33 & 36 \\ \hline
        Lassen wir die Welt nich allein & 30 & 40 \\ \hline
        Lauf deinem Traum nicht hinterher & 61 & 63 \\ \hline
        Liebe bis zum Tod & 37 & 26 \\ \hline
        Loyalität & 49 & 44 \\ \hline
        Luaa-rock'n Opposition & 64 & 78 \\ \hline
        Lügen und nette Märchen & 60 & 54 \\ \hline
        Mach das beste draus & 42 & 35 \\ \hline
        Mach dich auf & 58 & 55 \\ \hline
        Macht euch endlich alle platt & 61 & 82 \\ \hline
        Mal Heimweh, mal Fernweh & 45 & 55 \\ \hline
        Mal Sind mal Waren wir & 36 & 48 \\ \hline
        Medley still, unverzerrt und hart besaitet & 66 & 60 \\ \hline
        Mehr als eine Sünde & 70 & 60 \\ \hline
    \end{tabular}
    \caption{Rechtsrock probability (in \%) assigned to the individual \textit{Frei.Wild} songs by the classifiers. If the value was >50, they were classified as Rechtsrock. (continued)}
\end{table*}

\begin{table*}
\ContinuedFloat 
    \centering
    \begin{tabular}{|l|l|l|}
    \hline
        \textbf{\textit{Frei.Wild} song title} & \textbf{TF-IDF} & \textbf{sentence embedding} \\ \hline
        Mehr als tausend Worte & 45 & 37 \\ \hline
        Mein Lachen, dein Leiden & 37 & 45 \\ \hline
        Mein Leben, meine Geschichte, meine Lehre & 42 & 37 \\ \hline
        Meine Augen durch deine Augen & 40 & 40 \\ \hline
        Mensch oder Gott & 32 & 37 \\ \hline
        Mimimuttersöhnchen & 40 & 23 \\ \hline
        Miss America & 30 & 22 \\ \hline
        Mit dem Herz eines Adlers & 50 & 33 \\ \hline
        Montag ist ein Scheißtag & 41 & 42 \\ \hline
        Morgen wird alles besser & 41 & 31 \\ \hline
        Nehmen und nicht geben & 38 & 32 \\ \hline
        Nennt es Zufall nennt es Plan & 49 & 66 \\ \hline
        Nicht dein Tag & 44 & 52 \\ \hline
        Nicht zuviel denken und einfach machen & 53 & 46 \\ \hline
        Nichts kommt schlimmer als erwartet & 42 & 50 \\ \hline
        Niemand & 56 & 61 \\ \hline
        Nur das Leben & 45 & 51 \\ \hline
        Nur das Leben in Freiheit & 59 & 55 \\ \hline
        Nur Dumme sagen ja und amen & 66 & 75 \\ \hline
        Nur Gott richtet mich & 44 & 67 \\ \hline
        Nur Lieder die das Herz berühren & 60 & 44 \\ \hline
        Oben befiehlt, unten folgt & 73 & 84 \\ \hline
        Oft bekriegt nie besiegt & 54 & 43 \\ \hline
        Ohne dich kann ich nicht sein & 39 & 43 \\ \hline
        Planet voller Affen & 49 & 43 \\ \hline
        Rache muss sein & 37 & 45 \\ \hline
        Renne, brenne, Himmelstürmer & 47 & 48 \\ \hline
        Rookies and Kings & 20 & 48 \\ \hline
        Schau nach oben & 46 & 51 \\ \hline
        Scheiße für die Welt & 60 & 50 \\ \hline
        Schenkt uns Dummheit kein Niveau & 66 & 46 \\ \hline
        Schlagzeile groß Hirn zu klein & 62 & 55 \\ \hline
        Schlauer als der Rest & 75 & 72 \\ \hline
        Schmerz der Phantasie & 40 & 57 \\ \hline
        Schrei auf schrei laut & 23 & 48 \\ \hline
        Schwarz und weiss & 34 & 58 \\ \hline
        Schwarze Rosen & 51 & 60 \\ \hline
        Schwarzer Septemberregen & 55 & 86 \\ \hline
        Sei du dein Lieblingslied & 36 & 33 \\ \hline
        Sein oder nicht sein & 63 & 52 \\ \hline
        Selig oder Sünder & 50 & 48 \\ \hline
        Sie müssen es nicht wissen & 35 & 40 \\ \hline
        Sieg und Sorgen & 70 & 57 \\ \hline
        Sieger stehen da auf, wo Verlierer liegen bleiben & 41 & 60 \\ \hline
        Sieger stehen da wo verlierer liegen bleiben & 42 & 61 \\ \hline
    \end{tabular}
    \caption{Rechtsrock probability (in \%) assigned to the individual \textit{Frei.Wild} songs by the classifiers. If the value was >50, they were classified as Rechtsrock. (continued)}
\end{table*}

\begin{table*}
\ContinuedFloat 
    \centering
    \begin{tabular}{|l|l|l|}
    \hline
        \textbf{\textit{Frei.Wild} song title} & \textbf{TF-IDF} & \textbf{sentence embedding} \\ \hline
        Sommerland & 31 & 30 \\ \hline
        Sorgenleer & 35 & 28 \\ \hline
        Spirit of 96 & 30 & 29 \\ \hline
        Steine deiner Mauer & 40 & 47 \\ \hline
        Sternenstaub & 55 & 52 \\ \hline
        Stück für Stück & 54 & 48 \\ \hline
        Südtirol & 70 & 68 \\ \hline
        Terror & 60 & 67 \\ \hline
        The World Goes Down & 41 & 37 \\ \hline
        Tod und doch am Leben & 61 & 60 \\ \hline
        Tritt dir selber in den Arsch & 52 & 73 \\ \hline
        Trotzdem weitergehen & 50 & 50 \\ \hline
        Über Leichen gehen & 40 & 49 \\ \hline
        Unbrechbar & 55 & 33 \\ \hline
        Und dafür liebe ich dich & 25 & 24 \\ \hline
        Und ich war wieder da & 65 & 29 \\ \hline
        Unendliches Leben & 53 & 50 \\ \hline
        Unser Wille, unser Weg & 44 & 57 \\ \hline
        Unterwegs & 56 & 59 \\ \hline
        Unvergessen, Unvergänglich, Lebenslänglich & 61 & 42 \\ \hline
        Verbietet meine Freunde & 40 & 41 \\ \hline
        Verbotene Liebe, verbotene Kuss & 62 & 50 \\ \hline
        Verbrecher, Verlierer, Stalin und der Führer & 64 & 76 \\ \hline
        Verdammte Welt & 39 & 43 \\ \hline
        Vergangener Schmerz bricht kein gebrochenes Herz & 38 & 55 \\ \hline
        Vergiss mein nicht, vermisse mich & 56 & 43 \\ \hline
        Versteck dich oder bleib & 62 & 78 \\ \hline
        Völkerrecht & 65 & 62 \\ \hline
        Voll & 27 & 27 \\ \hline
        Volle Kanne & 33 & 23 \\ \hline
        Volle Pulle in die Fresse dieser Zeit & 56 & 70 \\ \hline
        Vom Regen in die Traufe & 57 & 41 \\ \hline
        Von der Wiege bis zur Bar & 23 & 20 \\ \hline
        Vorne liegt der Horizont & 39 & 62 \\ \hline
        Wahr oder gelogen & 31 & 33 \\ \hline
        Wahre Werte & 74 & 68 \\ \hline
        Warum? & 45 & 32 \\ \hline
        Was dann kommt werden wir sehen & 52 & 42 \\ \hline
        Was du liebst lass frei & 59 & 59 \\ \hline
        Weck mich auf & 27 & 34 \\ \hline
        Wecke deinen Helden auf & 46 & 38 \\ \hline
        Weder Gott weder die Hölle & 45 & 59 \\ \hline
        Weil du mich nur verarscht hast & 48 & 47 \\ \hline
        Weil ihr gerne Kriege führt & 78 & 77 \\ \hline
        Weil kein Krieg für ewig ist & 61 & 46 \\ \hline
    \end{tabular}
    \caption{Rechtsrock probability (in \%) assigned to the individual \textit{Frei.Wild} songs by the classifiers. If the value was >50, they were classified as Rechtsrock. (continued)}
\end{table*}

\begin{table*}
\ContinuedFloat 
    \centering
    \begin{tabular}{|l|l|l|}
    \hline
        \textbf{\textit{Frei.Wild} song title} & \textbf{TF-IDF} & \textbf{sentence embedding} \\ \hline
        Weiter immer weiter & 44 & 55 \\ \hline
        Wenn alles bricht & 44 & 56 \\ \hline
        Wenn alles in Trümmern liegt & 48 & 61 \\ \hline
        Wenn die Erinnerung erwacht & 48 & 27 \\ \hline
        Wenn mein Licht erlischt & 37 & 45 \\ \hline
        Wer nichts weiß wird alles glauben & 52 & 33 \\ \hline
        Wer weniger schläft ist länger wach & 32 & 32 \\ \hline
        Wer wenn nicht wir LUAA & 51 & 58 \\ \hline
        Wie Asche am Boden & 32 & 49 \\ \hline
        Wie ein schützender Engel & 45 & 39 \\ \hline
        Wie oft solln wir dir's noch sagen & 29 & 34 \\ \hline
        Willig, sexy und perfekt & 21 & 23 \\ \hline
        Wir brechen eure Seelen & 68 & 69 \\ \hline
        Wir bringen alle um & 41 & 64 \\ \hline
        Wir gegen alle & 57 & 58 \\ \hline
        Wir gehen dir ewig auf die Eier & 49 & 80 \\ \hline
        Wir gehen wie Bomben auf euch nieder & 49 & 75 \\ \hline
        Wir kentern nicht & 53 & 53 \\ \hline
        Wir reiten in den Untergang & 63 & 70 \\ \hline
        Wir sagen danke für all die ganzen Jahre & 76 & 60 \\ \hline
        Wir schaffen Deutsch.Land & 77 & 66 \\ \hline
        Wir sind viele & 53 & 51 \\ \hline
        Rivalen und Rebellen & 48 & 70 \\ \hline
        Wirklich so im Arsch & 37 & 28 \\ \hline
        Wo die Sonne wieder lacht & 53 & 34 \\ \hline
        Wo geht es hin wo bleiben wir stehen & 41 & 47 \\ \hline
        Wo nur die besten thronen & 35 & 57 \\ \hline
        Wochenendsparty & 42 & 30 \\ \hline
        Wünsche, Sehnsucht und Wille & 24 & 38 \\ \hline
        Yeah, yeah, yeah & 22 & 29 \\ \hline
        Zeig große Eier und ihnen den Arsch & 67 & 62 \\ \hline
        Zeig mir das Land & 43 & 42 \\ \hline
        Zerschlag dein Eis des Herzens & 29 & 63 \\ \hline
        Zieh mit den Göttern & 48 & 34 \\ \hline
        Ziel & 33 & 50 \\ \hline
        Zu hoch am Himmel & 42 & 46 \\ \hline
        Zufriedenheit & 24 & 26 \\ \hline
        Zusammen Freunde bleiben & 58 & 47 \\ \hline
        Zusammen und Vereint & 69 & 65 \\ \hline
        Zwischen allen Fronten & 63 & 55 \\ \hline
        Zwischen Trauer Liebe und Schmerz & 53 & 32 \\ \hline
    \end{tabular}
    \caption{Rechtsrock probability (in \%) assigned to the individual \textit{Frei.Wild} songs by the classifiers. If the value was >50, they were classified as Rechtsrock. (continued)}
\end{table*}

\end{document}